# From Single Scan to Sequential Consistency: A New Paradigm for LiDAR Relocalization

Minghang Zhu[1,2*], Zhijing Wang[1,2*], Yuxin Guo[1,2], Wen Li[1,2], Sheng Ao[1,2†], Cheng Wang[1,2†]
Fujian Key Laboratory of Sensing and Computing for Smart Cities, Xiamen University
Ministry of Education of China, Xiamen University

## Abstract

*LiDAR relocalization aims to estimate the global 6-DoF pose of a sensor in the environment. However, existing regression-based approaches are prone to dynamic or ambiguous scenarios, as they either solely rely on single-frame inference or neglect the spatio-temporal consistency across scans. In this paper, we propose **TempLoc**, a new LiDAR relocalization framework that enhances the robustness of localization by effectively modeling sequential consistency. Specifically, a Global Coordinate Estimation module is first introduced to predict point-wise global coordinates and associated uncertainties for each LiDAR scan. A Prior Coordinate Generation module is then presented to estimate inter-frame point correspondences by the attention mechanism. Lastly, an Uncertainty-Guided Coordinate Fusion module is deployed to integrate both predictions of point correspondence in an end-to-end fashion, yielding a more temporally consistent and accurate global 6-DoF pose. Experimental results on the NCLT and Oxford Robot-Car benchmarks show that our TempLoc outperforms state-of-the-art methods by a large margin, demonstrating the effectiveness of temporal-aware correspondence modeling in LiDAR relocalization. Our code will be released soon.*

## 1. Introduction

Accurate and robust LiDAR-based relocalization is a fundamental capability for autonomous systems [13, 28, 58] and robotics [37, 43, 47]. Given a pre-built 3D map, the goal of LiDAR-based localization is to estimate the global 6 degree of freedom (DoF) pose of sensor according to the captured real-time scans [1, 3, 36, 59]. The primary difficulty lies in handling an environment with dynamic variations, occlusions, and sensor noise [2, 12, 54].

Most existing LiDAR relocalization methods [10, 21] rely on a matching paradigm, where the input point cloud is aligned against a pre-built global map to estimate the current pose [26, 31]. While highly accurate, these meth-ods require storing and querying large-scale 3D maps, resulting in significant memory and communication overhead [17, 33, 38]. In contrast, recent regression-based methods [20, 45] attempt to directly predict the absolute pose from a single point cloud scan using deep neural networks, bypassing the need for explicit map storage and achieving higher computational efficiency. These regression-based alternatives open up promising directions for deployment on resource-constrained platforms [53].

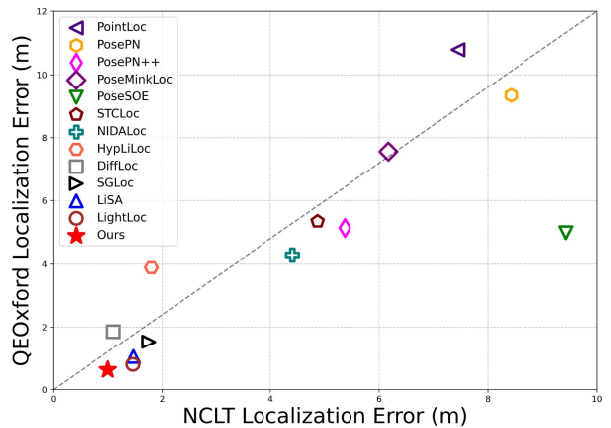

Figure 1. **Localization Error on Different Datasets.** Experiments are conducted on the QEOxford and NCLT datasets, where closer proximity to 0 on the x-axis and y-axis indicates better localization performance. The proposed TempLoc achieves the best performance on both datasets, with significant improvements against the state-of-the-art methods.

Nevertheless, the regression-based methods often struggle in complex or ambiguous environments [5–7]. This is because single LiDAR scan is prone to measurement noise and observable geometric variations induced by spatial changes or dynamic objects. To alleviate this issue, a handful of work [56, 57] has been proposed by incorporating LiDAR sequences to enhance localization accuracy. However, they only roughly encode spatial-temporal global features, without explicitly modeling the point-level sequential consistency [16], which results in suboptimal localization performance.

In the 2D counterpart, recent advances in temporal relocalization have demonstrated that incorporating frame-to-frame correspondences at the pixel level can significantly improve performance [14, 49]. For instance, KFNet [60] estimates per-pixel scene coordinates and model the temporal transition using optical flow, enabling recursive state estimation through Kalman filtering. While effective for 2D image, these approaches [11, 15, 50] cannot be directly extended to the 3D domain due to the unordered and irregular nature of point clouds in 3D scenes, coupled with challenges such as sparsity, occlusions, and interference from dynamic objects, temporal LiDAR-based localization faces unique difficulties.

This observation raises a critical question: can a similar paradigm be adapted to LiDAR relocalization? Motivated by this, we propose a new LiDAR relocalization framework that jointly performs measurement and prior estimation, and subsequently integrates them through a learned fusion mechanism. Our method, named TempLoc, consists of three components: (1) the **Global Coordinate Estimation** that predicts per-point global coordinates and uncertainties from a single LiDAR scan, (2) the **Prior Coordinate Generation** that leverages a point cloud registration network (*i.e.,* PCAM [8]) to estimate inter-frame point-wise transitions, and (3) the **Uncertainty-Guided Coordinate Fusion** that employs an end-to-end differentiable fusion mechanism to generate accurate and temporally consistent point correspondences. In the end, the global 6-DoF pose is obtained via a RANSAC-based optimization using the refined 3D correspondences.

Benefiting from explicitly modeling the temporal evolution of scene coordinates, our method effectively mitigates the limitations of single-frame regression and significantly enhances localization robustness. As shown in Fig. 1, our method achieves state-of-the-art performance on three public available benchmarks. Notably, the localization accuracy of our method outperforms that of the strong baseline LightLoc [25] by nearly 30% on the NCLT [9] dataset. Overall, our contributions are three-fold:

- We propose a temporally-aware LiDAR relocalization framework that incorporates sequential constraints into point correspondence estimation by extending scene coordinate regression to the temporal domain.

- We introduce an uncertainty-aware coordinate estimation module to predicts per-point global coordinates from single LiDAR scan.

- Extensive experiments and ablations demonstrating the effectiveness of our method and providing the intuition for its architectural components.

## 2. Related Work

Unlike map-based approaches [22, 40], regression-based solutions directly predict the global pose. These methods can be further categorized into Absolute Pose Regression (APR) and Scene Coordinate Regression (SCR).

### 2.1. Absolute Pose Regression

Absolute pose regression aims to use deep networks to learn and memorize scene information, directly regressing the sensor's global pose in an end-to-end manner. PoseNet [19] is a classic solution of APR, which proposes to use a modified GoogleNet [39] to regress camera poses. The first proposed APR method for LiDAR is PointLoc [46], which uses PointNet++ [32] and a self-attention module to extract global features. PosePN [55] adopts a universal encoder and memory-aware regression to avoid redundant retraining and improve localization performance. HypLiLoc [44] combines 3D features and 2D spherical projection features to further improve the performance. DiffLoc [24] introduces a diffusion model to obtain robust and accurate positioning through an iterative denoising process. These methods use single-frame data as input information, which will result in many outliers.

Since sequential data can provide more contextual information for the model and introduce time constraints, VLocNet [41] proposes to combine relative pose regression with APR and adopt a multi-task alternating optimization strategy to learn shared features in the entire network. Its variant VLocNet++ [34] further aggregates semantic information into the model to learn discriminative features. LSG [48] introduced a visual odometry component as a local motion constraint to alleviate the uncertainty of pose estimation. ViPR [30] effectively integrates APR and relative pose estimation using LSTMs, leveraging their complementary strengths. CoordiNet [29] jointly trains pose predictions and uncertainties, and eliminates positioning outliers based on Kalman filtering. STCLoc [56] and NIDALoc [57] are two LiDAR-specific schemes that use sequence data as input. STCLoc introduces spatial and temporal constraints to effectively regularize absolute pose regression, and NIDALoc designs a memory module to use historical information to reduce scene ambiguity. However, existing LiDAR-based APR methods rely on long-term sequence information, which leads to a substantial computational load and affects their real-time capabilities.

### 2.2. Scene Coordinate Regression

Different from APR which directly regresses the pose, SCR regresses the point cloud coordinates in the world coordinate system and relies on RANSAC for pose estimation. SGLoc [23] applied SCR to LiDAR positioning for the first time and significantly improved the positioning performance. LiSA [51] applies diffusion-based knowledge dis-tillation to enable the SCR network to have semantic understanding and focus on more important points. In order to reduce training time, LightLoc [25] proposed a universal encoder that compresses training time to 1 hour without losing model performance.

However, Scene Coordinate Regression (SCR) often yields numerous outliers in regressed point coordinates. Despite iterative denoising with RANSAC, these outliers can still compromise localization accuracy. To mitigate the impact of such outliers, we integrate uncertainty estimation. Currently, within a filter framework, we fuse temporal information from neighboring frames to achieve superior pose estimation and smoother trajectories.

## 3. Methodology

As shown in Fig. 2, our method consists of three key modules: Global Coordinate Estimation (GCE), Prior Coordinate Generation (PCG), and Uncertainty-guided Coordinate Fusion (UCF). The GCE module produces 3D point correspondences for each LiDAR scan between the local and global coordinate systems, along with their associated uncertainty scores. Simultaneously, the PCG module estimates 3D point correspondences between adjacent frames and provides corresponding credibility assessments. Based on this, the UCF module integrates the outputs of the GCE and PCG to obtain refined 3D-3D correspondences between the local and global coordinate systems. Finally, the global 6-DoF pose is computed through RANSAC.

### 3.1. Global Coordinate Estimation

The SCR method is prone to producing inaccurate coordinate predictions in environments characterized by noise, dynamic disturbances, or insufficient structural information. However, existing LiDAR relocalization methods typically overlook the quality of per-point predictions. To this end, we incorporate a per-point uncertainty estimation mechanism into the SCR architecture to assess the quality of each predicted global coordinate.

**Prediction of uncertainty.** For an input point cloud $\mathbf{P} \in \mathbb{R}^{N \times 3}$, we utilize the LightLoc network [25] to regress scene coordinates $\mathbf{P}_{pred} \in \mathbb{R}^{M \times 3}$ in global reference frame. Simultaneously, the per-point uncertainty score $\boldsymbol{u}_{pred} \in \mathbb{R}^{M \times 1}$ can be predicted by shared MLPs. Note that, the ground-truth score $u_{gt}$ is defined as:

$$u_i^{gt} = \begin{cases} 0, & \sum_{i=1}^{M} \left\| \boldsymbol{p}_i^{pred} - \boldsymbol{p}_i^{gt} \right\|_1 < \tau, \\ 1, & \text{otherwise.} \end{cases} \quad (1)$$

where $\mathbf{P}^{gt}$ represents the ground-truth global coordinate of each point. The threshold $\tau$ is not static and decays as the network training progresses.

This strategy is critical: in the early training stages, when prediction errors are large, a higher $\tau$ allows the network to treat a broad range of predictions as confident, focusing its learning on identifying and penalizing the severe outliers. This prevents a trivial learning signal where nearly all points are labeled as high-uncertainty. As the network converges and its predictions become more accurate, a progressively smaller $\tau$ tightens the criterion, compelling the model to learn finer-grained distinctions where only the most precise predictions earn a low-uncertainty label, thereby refining its uncertainty estimation capability.

**Loss Function.** The total loss is composed of a regression loss and an uncertainty loss. The uncertainty prediction is supervised by a dedicated loss term, $\mathcal{L}_{un}$. We employ the Mean Squared Error (MSE) loss, which computes the average of the squared differences between the predicted uncertainty $u_i^{pred}$ and the ground truth label $u_i^{gt}$:

$$\mathcal{L}_{un} = \frac{1}{M} \sum_{i=1}^{M} (u_i^{pred} - u_i^{gt})^2. \quad (2)$$

Meanwhile, the primary regression objective, $\mathcal{L}_{reg}$, minimizes the mean distance L1 [35] between the predicted and ground truth coordinates:

$$\mathcal{L}_{reg} = \frac{1}{M} \sum_{i=1}^{M} \left\| \boldsymbol{p}_i^{pred} - \boldsymbol{p}_i^{gt} \right\|_1. \quad (3)$$

The final loss $\mathcal{L}_{GCE}$ is the sum of these two terms:

$$\mathcal{L}_{GCE} = \mathcal{L}_{reg} + \mathcal{L}_{un}. \quad (4)$$

### 3.2. Prior Coordinate Generation

To effectively leverage temporal information, we design a prior coordinate generation module to generate a reliable world coordinate prior for the current frame. This process has two core stages: first, generating a soft correspondence and uncertainty in the current frame for each point in the previous frame using a registration network; second, using these correspondences to assign prior coordinates to points in the current frame through a neighborhood interpolation.

**Soft Correspondence Generation.** We adopt the PCAM [8] as the backbone network and additionally introduce a self-attention mechanism [42] to enhance point-wise features within each point cloud, thereby generating discriminative feature descriptors.

For a point cloud $\mathbf{P}^{(t)}$ with initial features $\mathbf{F}^{(t)}$, we first compute its self-attention matrix $\mathbf{A}^{(t,t)}$. Each element $(\mathbf{A}^{(t,t)})_{ij}$, representing the attention of point $i$ on point $j$, is

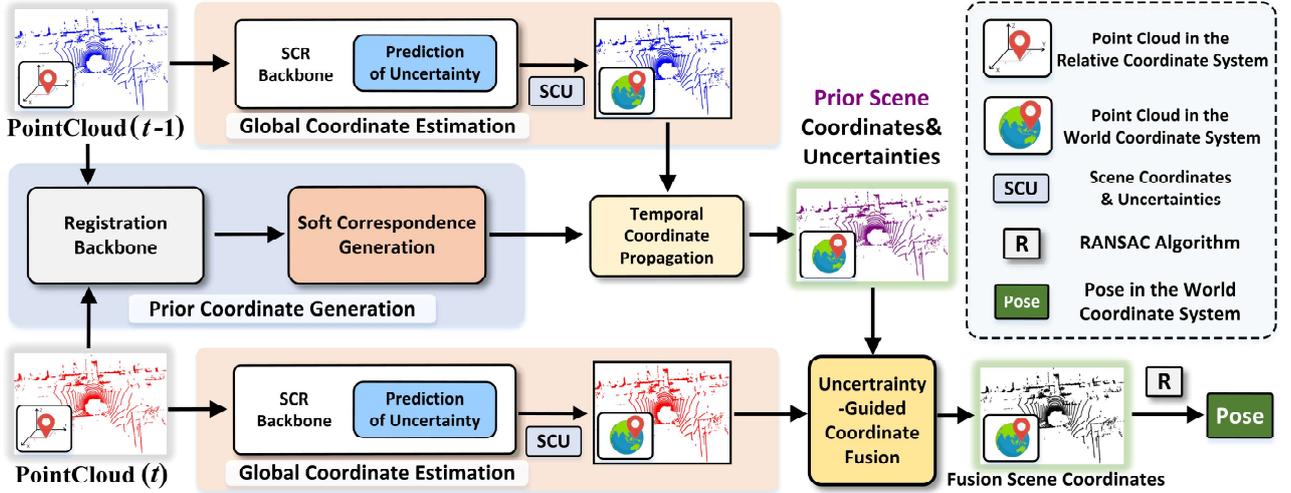

Figure 2. **Overview of the proposed framework.** Temporal point clouds from consecutive frames are processed through a SCR module with uncertainty estimation to generate stable pose estimates. Subsequently, a prior registration module refines the pose estimates using soft correspondences. Then, uncertainty-guided coordinate fusion produces point clouds in the world coordinate system. Finally, the RANSAC algorithm is employed to solve for the point cloud localization pose.

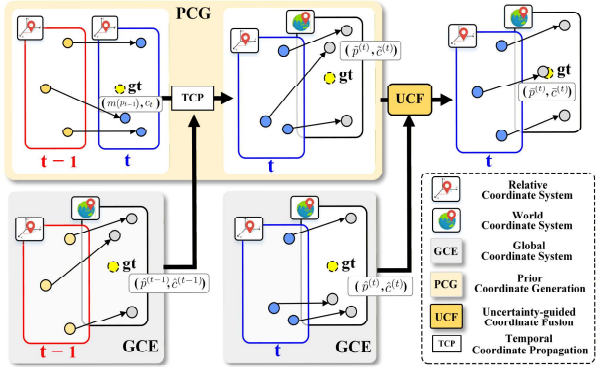

Figure 3. **Illustration of coordinate fusion.** The GCE module produces the measurement estimation at $t$ time, while the PCG module provides the prior estimation at the same time. These two estimations are fused by the UCF module to generate more accurate point correspondences between the relative and world coordinate systems.

computed from the cosine similarity of intra-cloud feature pairs, followed by a Softmax normalization:

$$s_{ij} = \frac{\mathbf{F}_i^{(t)} \cdot (\mathbf{F}_j^{(t)})^T}{||\mathbf{F}_i^{(t)}||_2 \cdot ||\mathbf{F}_j^{(t)}||_2}, \quad (5)$$

$$(\mathbf{A}^{(t,t)})_{ij} = \frac{e^{s_{ij}}}{\sum_{k=1}^N e^{s_{ik}}}. \quad (6)$$

After obtaining the context-aware features, we use them to compute multi-level cross-attention from current frame $\mathbf{P}^{(t)}$ to previous frame $\mathbf{P}^{(t-1)}$. We aggregate all $L$ cross-attention matrices $\mathbf{A}_{(l)}^{(t-1,t)}$ through element-wise multiplication to obtain a global attention matrix $\mathbf{A}_{global}^{(t-1,t)}$:

$$\mathbf{A}_{global}^{(t-1,t)} = \mathbf{A}_{(1)}^{(t-1,t)} \odot \mathbf{A}_{(2)}^{(t-1,t)} \odot \cdots \odot \mathbf{A}_{(L)}^{(t-1,t)}. \quad (7)$$

Based on this global attention matrix, we generate a soft correspondence $m(\boldsymbol{p}_i^{(t-1)})$ in the current frame for each point $\boldsymbol{p}_i^{(t-1)}$ from the previous frame. This soft correspondence is a weighted average of all points in the current frame, with weights given by the corresponding row vector of the global attention matrix:

$$m(\boldsymbol{p}_i^{(t-1)}) = \frac{\sum_{j=1}^N (\mathbf{A}_{global}^{(t-1,t)})_{ij} \cdot \boldsymbol{p}_j^{(t)}}{\sum_{k=1}^N (\mathbf{A}_{global}^{(t-1,t)})_{ik}}, \quad (8)$$

where $\boldsymbol{p}_j^{(t)}$ is the point in current frame cloud $\mathbf{P}^{(t)}$, and $N$ is its number of points. This yields a set of pseudo-correspondence points $\{m(\boldsymbol{p}_i^{(t-1)})\}$, where each point carries the temporal information of its origin point $\boldsymbol{p}_i^{(t-1)}$.

**Temporal Coordinate Propagation.** The generated soft correspondence point $m(\boldsymbol{p}_i^{(t-1)})$ is a theoretical position and does not necessarily match any of the actual sampled points $\boldsymbol{p}_j^{(t)}$ in the current frame. To assign a prior coordinate to the actual points of the current frame, we use a neighborhood interpolation strategy.

Assume we have predicted the world coordinates of all points in the previous frame, denoted as $\{\hat{\boldsymbol{p}}_i^{(t-1)}\}$. For each actual point $\boldsymbol{p}_j^{(t)}$ in the current frame, we search for its $k$ nearest neighbors within the pseudo-correspondence set $\{m(\boldsymbol{p}_i^{(t-1)})\}$. Let these $k$ nearest pseudo-neighbors be $\{m(\boldsymbol{p}_{m_1}^{(t-1)}), \ldots, m(\boldsymbol{p}_{m_k}^{(t-1)})\}$. Since each pseudo-correspondence point $m(\boldsymbol{p}_m^{(t-1)})$ corresponds to a source point $\boldsymbol{p}_m^{(t-1)}$ from the previous frame, it can inherit the corresponding world coordinate $\hat{\boldsymbol{p}}_m^{(t-1)}$. Next, we compute the prior world coordinate $\tilde{\boldsymbol{p}}_j^{(t)}$ for the actual current-frame point $\boldsymbol{p}_j^{(t)}$ by performing a distance-weighted average of the world coordinates carried by these $k$ pseudo-neighbors:

$$\tilde{\boldsymbol{p}}_j^{(t)} = \sum_{l=1}^{k} w_{j,l} \cdot \hat{\boldsymbol{p}}_{m_l}^{(t-1)}, \tag{9}$$

the weight is defined as $w_{j,l} = \frac{1/d_{j,l}}{\sum_{s=1}^{k} 1/d_{j,s}}$, where $d_{j,l}$ represents the Euclidean distance from the actual point $\boldsymbol{p}_j^{(t)}$ to its $l$-th pseudo-neighbor $m(\boldsymbol{p}_{m_l}^{(t-1)})$.

**Loss Function.** The propagated prior coordinates $\tilde{\mathbf{P}}^{(t)}$ and their associated uncertainties are jointly supervised by the loss $\mathcal{L}_{PCG}$, which is analogous to the definitions in Equations (2)-(4).

### 3.3. Uncertainty-Guided Coordinate Fusion

For each LiDAR scan, we can obtain two groups of world coordinates: (1) the *measurement* estimation $(\hat{\mathbf{P}}^{(t)}, \hat{u}^{(t)})$ comprising the coordinates and uncertainties directly predicted by the Global Coordinate Estimation, and (2) the *prior* estimation $(\tilde{\mathbf{P}}^{(t)}, \tilde{u}^{(t)})$ obtained from the Prior Coordinate Generation. To obtain the final and more accurate coordinate estimation, we design the Uncertainty-Guided Coordinate Fusion module, as shown in Fig. 3.

Inspired by the Kalman filtering [18], which involves an adaptive weighted fusion based on the uncertainty of different sources, we design a data-driven and end-to-end differentiable fusion mechanism. Specifically, for each point, we concatenate its prior uncertainty $\tilde{u}_i^{(t)}$ and measurement uncertainty $\hat{u}_i^{(t)}$ and use a Softmax function to dynamically compute the fusion weights. The Softmax function ensures the weights are positive and sum to one, which provides a probabilistic interpretation for the fusion, and allowing the network to automatically learn how to balance the two sources based on data:

$$[\alpha_i, \beta_i] = \text{Softmax}([-\tilde{u}_i^{(t)}, -\hat{u}_i^{(t)}]), \tag{10}$$

where $\alpha_i$ is the weight assigned to the prior coordinate $\tilde{\boldsymbol{p}}_i^{(t)}$ and $\beta_i$ is the weight for the measurement coordinate $\hat{\boldsymbol{p}}_i^{(t)}$. The final fused coordinates $\bar{\boldsymbol{p}}^{(t)}$ and uncertainty $\bar{u}^{(t)}$ are then computed by weighted summation:

$$\bar{\boldsymbol{p}}_i^{(t)} = \alpha_i \tilde{\boldsymbol{p}}_i^{(t)} + \beta_i \hat{\boldsymbol{p}}_i^{(t)}, \tag{11}$$

$$\bar{u}_i^{(t)} = \alpha_i \tilde{u}_i^{(t)} + \beta_i \hat{u}_i^{(t)}. \tag{12}$$

The weighted outputs are applied to both the coordinates and their associated uncertainty scores, resulting in the fused estimations with consistence and reliability.

**Loss Function.** During the training phase, the fused results are supervised by the loss $\mathcal{L}_{Fuse}$, which is analogous to the losses in Equations (2)-(4). Finally, the overall loss of our framework can be formulated as:

$$\mathcal{L}_{full} = \lambda_1 \mathcal{L}_{GCE} + \lambda_2 \mathcal{L}_{PCG} + \lambda_3 \mathcal{L}_{Fuse}, \tag{13}$$

where $\lambda_1$, $\lambda_2$, and $\lambda_3$ are set to 0.3, 0.3, and 0.4, respectively, and control the weights of the three losses.

## 4. Experiments

In this section, we first describe the experimental setup, including the datasets and implementation details (Sec. 4.1). We then conduct extensive experiments to compare the proposed TempLoc with the state-of-the-art methods on typical benchmarks (Sec. 4.2). Lastly, a series of ablation studies are conducted (Sec. 4.3).

### 4.1. Experimental Setup

**Benchmark Datasets.** We conduct evaluation experiments on two widely-used benchmark datasets for outdoor LiDAR-based localization: the Oxford RobotCar Dataset [4] and the NCLT Dataset [9]. During evaluation, unified localization metrics are adopted: average translation error and average rotation error.

**Oxford RobotCar Dataset.** This dataset was collected by repeatedly traversing a fixed route in central Oxford, UK, using a Velodyne-32 LiDAR sensor, with ground truth obtained from GPS and INS. Each route is approximately 10 km in length. Consistent with other localization methods, we use four routes for training (11-14-02-26, 14-12-05-52, 14-14-48-55,18-15-20-12) and four routes for testing (15-13-06-37, 17-13-26-39, 17-14-03-00, 18-14-14-42). Additionally, we follow SGLoc [23] to construct the QEOxford dataset for evaluation, which is an enhanced version of the Oxford dataset using ground-truth corrections through alignment techniques.

**NCLT Dataset.** This dataset was collected at the University of Michigan's North Campus using a Segway robotic platform equipped with a Velodyne-32 LiDAR sensor. It encompasses both indoor and outdoor environments with seasonal variations. Ground truth is obtained using GPS refined by SLAM techniques. The average route length is approximately $5.5km$. Following SGLoc [23], we select four routes for training (2012-01-22, 2012-02-02, 2012-02-18, 2012-05-11) and four routes for testing (2012-02-12, 2012-02-19, 2012-03-31, 2012-05-26).

Table 1. **Average translation error(m) and rotation error(°) on the QEOxford dataset.** The best results are indicated in **bold**, and the second-best results are underlined.

|     | Methods | 15-13-06-37 | 17-13-26-39 | 17-14-03-00 | 18-14-14-42 | Average[m/°] |
|---|---|---|---|---|---|---|
| APR | PointLoc [46] | 10.75/2.36 | 11.07/2.21 | 11.53/1.92 | 9.82/2.07 | 10.79/2.14 |
|     | PosePN++ [32] | 4.54/1.83 | 6.44/1.78 | 4.89/1.55 | 4.64/1.61 | 5.13/1.69 |
|     | PoseSOE [55] | 4.17/1.76 | 6.16/1.81 | 5.42/1.87 | 4.16/1.70 | 4.98/1.79 |
|     | STCLoc [56] | 5.14/1.27 | 6.12/1.21 | 5.32/1.08 | 4.76/1.19 | 5.34/1.19 |
|     | NIDALoc [57] | 3.71/1.50 | 5.40/1.40 | 3.94/1.30 | 4.08/1.30 | 4.28/1.38 |
|     | HypLiLoc [44] | 5.03/1.46 | 4.31/1.43 | 3.61/1.11 | 2.61/1.09 | 3.89/1.27 |
|     | DiffLoc [24] | 2.03/<u>1.04</u> | 1.78/**0.79** | 2.05/**0.83** | 1.56/**0.83** | 1.86/**0.87** |
| SCR | SGLoc [23] | 1.79/1.67 | 1.81/1.76 | 1.33/1.59 | 1.19/1.39 | 1.53/1.60 |
|     | LiSA [51] | 0.94/1.10 | 1.17/1.21 | 0.84/1.15 | 0.85/1.11 | 0.95/1.14 |
|     | LightLoc [25] | <u>0.82</u>/1.12 | <u>0.85</u>/1.07 | <u>0.81</u>/1.11 | <u>0.82</u>/1.16 | <u>0.83</u>/1.12 |
|     | **TempLoc** | **0.74/0.99** | **0.77**/<u>0.99</u> | **0.60**/<u>0.95</u> | **0.77**/<u>1.05</u> | **0.72**/<u>0.99</u> |

Table 2. **Average translation error(m) and rotation error (°) on the Oxford dataset.** The best result sare indicated in **bold**, and the second-best results are underlined.

|     | Methods | 15-13-06-37 | 17-13-26-39 | 17-14-03-00 | 18-14-14-42 | Average[m/°] |
|---|---|---|---|---|---|---|
| APR | PointLoc [46] | 12.42/2.26 | 13.14/2.50 | 12.91/1.92 | 11.31/1.98 | 12.45/2.17 |
|     | PosePN++ [32] | 9.59/1.92 | 10.66/1.92 | 9.01/1.51 | 8.44/1.71 | 9.43/1.77 |
|     | PoseSOE [55] | 7.59/1.94 | 10.39/2.08 | 9.21/2.12 | 7.27/1.87 | 8.62/2.00 |
|     | STCLoc [56] | 6.93/1.48 | 7.55/1.23 | 7.44/1.24 | 6.13/1.15 | 7.01/1.28 |
|     | NIDALoc [57] | 5.45/1.40 | 7.63/1.56 | 6.68/1.26 | 4.80/1.18 | 6.14/1.35 |
|     | HypLiLoc [44] | 6.88/1.09 | 6.79/1.29 | 5.82/<u>0.97</u> | 3.45/<u>0.84</u> | 5.74/<u>1.05</u> |
|     | DiffLoc [24] | 3.57/**0.88** | 3.65/**0.68** | 4.03/**0.70** | 2.86/**0.60** | 3.53/**0.72** |
| SCR | SGLoc [23] | 3.01/1.91 | 4.07/2.07 | 3.37/1.89 | 2.12/1.66 | 3.14/1.88 |
|     | LiSA [51] | 2.36/1.29 | 3.47/1.43 | 3.19/1.34 | **1.95**/1.23 | 2.74/1.32 |
|     | LightLoc [25] | <u>2.33</u>/1.21 | <u>3.19</u>/1.34 | <u>3.11</u>/1.24 | 2.05/1.20 | <u>2.67</u>/1.25 |
|     | RALoc [52] | 3.19/4.10 | 3.87/3.96 | 3.32/3.87 | 2.59/3.71 | 3.24/3.91 |
|     | **TempLoc** | **2.22**/<u>1.05</u> | **3.08**/<u>1.12</u> | **2.94**/1.06 | <u>1.99</u>/1.07 | **2.55**/1.07 |

**Training Details** The proposed TempLoc method is implemented with PyTorch. To ensure fairness, the comparison methods utilize the officially provided code and pre-trained models. During training, the batch size is set to 64, and the Adam optimizer is used with an initial learning rate of 0.001. For the Oxford dataset, the point cloud is sampled with a voxel size of 0.25, while for the NCLT dataset, the voxel size is set to 0.3. The parameter $\tau$ is related to learning rate decay, we reduce $\tau$ by ×0.7 every 6 epochs. All experiments are conducted on the platform with Intel Xeon CPU@2.30GHZ with four NVIDIA RTX 3090Ti GPUs. Additional details are provided in the supplementary material.

**Baselines.** The proposed TempLoc method is compared with several state-of-the-arts, including both APR and SCR methods. Specifically, the APR baselines contain single-frame methods such as PointLoc [46], PosePN++ [55], PoseSOE [55], and HypLiLoc [44], as well as temporal approaches like STCLoc [56], NIDALoc [57], and DiffLoc [24]. For SCR methods, we consider SGLoc [23], LiSA [51], LightLoc [25], and RALoc [52]. RALoc is a recent work whose code is not yet released, so we directly report its results on Oxford. To verify the advantages of our map-free method over retrieval-based localization, we also compare our approach with BEVplace++ [27].

### 4.2. Evaluation

**Evaluation on the Oxford Dataset.** We first evaluate the proposed TempLoc on the QEOxford. As shown in Tab. 1, the proposed TempLoc method outperforms all existing localization methods in terms of average localization accuracy. Compared to the best-performing LightLoc, TempLoc reduces the translation error from $0.83m$ to $0.72m$ (a 13% improvement) and the rotation error from $1.12°$ to $0.99°$ (a 12% improvement). Although DiffLoc obtains superior orientation accuracy, its reliance on a sequence of three or more temporally separated frames. Our TempLoc requires

Table 3. **Average translation error(m) and rotation error(°) on the NCLT dataset.** The best result sare indicated in **bold**, and the second-best results are underlined. † denotes the removal of certain erroneous test segments following the LightLoc.

|     | Methods      | 2012-02-12 | 2012-02-19 | 2012-03-31 | 2012-05-26† | Average[m/°] |
|-----|--------------|------------|------------|------------|-------------|--------------|
| APR | PointLoc [46]   | 7.23/4.88  | 6.31/3.89  | 6.71/4.32  | 9.55/5.21   | 7.45/4.58    |
|     | PosePN++ [32]   | 4.97/3.75  | 3.68/2.65  | 4.35/3.38  | 8.42/4.30   | 5.36/3.52    |
|     | PoseSOE [55]    | 13.09/8.05 | 6.16/4.51  | 5.24/4.56  | 13.27/7.85  | 9.44/6.24    |
|     | STCLoc [56]     | 4.91/4.34  | 3.25/3.10  | 3.75/4.04  | 7.53/4.95   | 4.86/4.11    |
|     | NIDALoc [57]    | 4.48/3.59  | 3.14/2.52  | 3.67/3.46  | 6.32/4.67   | 4.40/3.56    |
|     | HypLiLoc [44]   | 1.71/3.56  | 1.68/2.69  | 1.52/2.90  | 2.29/3.34   | 1.80/3.12    |
|     | DiffLoc [24]    | 0.99/<u>2.40</u> | 0.92/<u>2.14</u> | 0.98/<u>2.27</u> | **1.36/2.48** | <u>1.06</u>/<u>2.32</u> |
| SCR | SGLoc [23]      | 1.20/3.08  | 1.20/3.05  | 1.12/3.28  | 3.48/4.43   | 1.75/3.46    |
|     | LiSA [51]       | <u>0.97</u>/**2.23** | 0.91/**2.09** | 0.87/**2.21** | 3.11/<u>2.72</u> | 1.47/**2.31** |
|     | LightLoc [25]   | 0.98/2.76  | <u>0.89</u>/2.51 | <u>0.86</u>/2.67 | 3.10/3.26 | 1.46/2.80 |
|     | **TempLoc**     | **0.74**/2.48 | **0.74**/2.22 | **0.68**/2.35 | <u>2.04</u>/3.20 | **1.05**/2.56 |

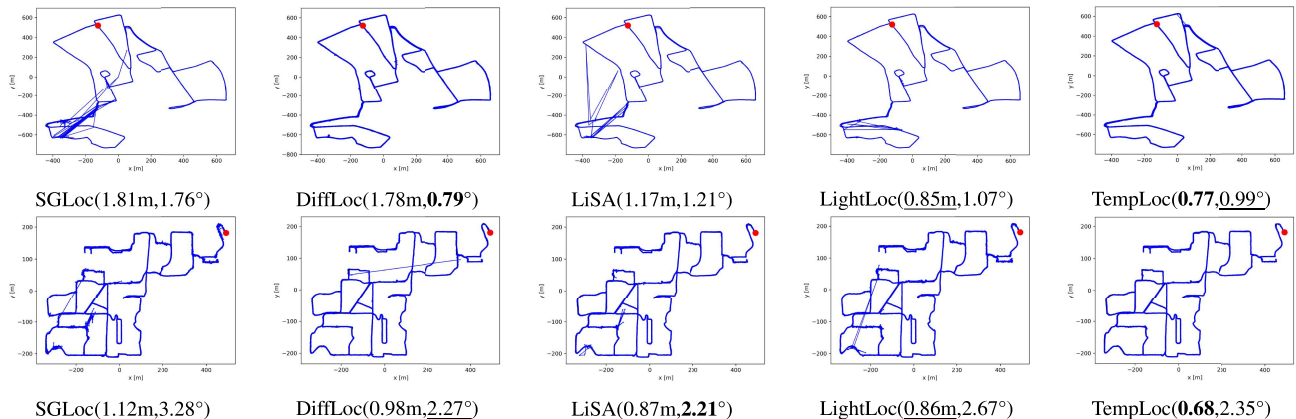

SGLoc(1.81m,1.76°)　　DiffLoc(1.78m,**0.79°**)　　LiSA(1.17m,1.21°)　　LightLoc(<u>0.85m</u>,1.07°)　　TempLoc(**0.77**,<u>0.99°</u>)

SGLoc(1.12m,3.28°)　　DiffLoc(0.98m,<u>2.27°</u>)　　LiSA(0.87m,**2.21°**)　　LightLoc(<u>0.86m</u>,2.67°)　　TempLoc(**0.68**,2.35°)

Figure 4. **Localization Results Visualization.** The first and second rows display the localization results on the QEOxford and NCLT datasets, respectively. Blue trajectories represent our predicted results, while black lines denote the ground truth. The best results are indicated in **bold**, and the second-best results are underlined.

only two consecutive point clouds and achieves a 61.3% improvement in localization accuracy, demonstrating a significant performance gain with a minimal temporal dependency. Additionally, DiffLoc exhibits significantly lower translation error compared to our TempLoc (with a 61.2% improvement), indicating that TempLoc is more robust in localization.

The top row of Fig. 4 shows the localization results for the QEOxford dataset. SGLoc and LiSA exhibit noticeable localization fluctuations, resulting in higher localization and angular errors compared to other methods. DiffLoc, leveraging temporal constraints, produces smoother trajectories with smaller angular errors but suffers from significant overall trajectory offsets, leading to high localization errors. LightLoc, as a high-performing method, still shows minor fluctuations. In contrast, the localization trajectories generated by the proposed TempLoc are smoother.

Due to QEOxford's adoption of global trajectory alignment from SGLoc [23], the ground truth error in the Oxford dataset was corrected. To ensure experimental fairness, we re-conducted experiments on the Oxford localization baseline, as shown in Tab. 2. It can be observed that all localization methods experienced a noticeable decline in localization accuracy on the Oxford dataset. Nevertheless, our TempLoc still achieved the best performance in terms of average localization accuracy.

**Evaluation on the NCLT Dataset.** As shown in Tab. 3, TempLoc was evaluated on four test trajectories, achieving an average localization accuracy improvement of 28% over LightLoc. When compared with DiffLoc, its accuracy on the fourth trajectory was notably lower. This is attributed to the absence of road segments in the test scene; DiffLoc leverages the powerful generative capabilities of its Diffusion Model to complete and infer missing structural information, whereas our baseline, LightLoc, lacks a similar capability, which in turn affects our algorithm's performance. However, for the first three trajectories, TempLoc's average localization accuracy surpassed DiffLoc's by approximately 25%, validating the effectiveness of our architecture in en-

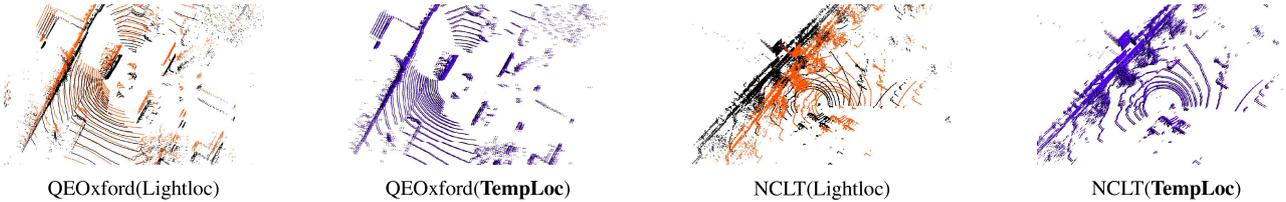

QEOxford(Lightloc)    QEOxford(**TempLoc**)    NCLT(Lightloc)    NCLT(**TempLoc**)

Figure 5. **localization visualization comparison.** Experiments were conducted on the QEOxford and NCLT datasets. Black point clouds represent the ground truth, **orange** points denote predictions from the baseline, and **BlueViolet** indicate predictions from our TempLoc.

Table 4. **Real-time, Overhead and Performance on NCLT.**

| Method | Real-time (↓) | Test GPU (↓) | Storage(↓) Model+Map | Error(↓) [m/°] | Recall@1 <5m(↓) |
|---|---|---|---|---|---|
| LightLoc [25] | 48ms | 2.1GB | 70MB+0MB | 3.10/3.26 | 96.5% |
| BEVplace++ [27] | 54ms | 1.95GB | 17MB+**841**MB | 11.83/4.64 | 92.7% |
| **TempLoc(Ours)** | 68ms | 2.3GB | 77MB+0MB | **2.04/3.20** | **98.3%** |

Table 5. **Ablation study on different modules.** **SCR+Conf** denotes applying uncertainty estimation on the backbone network, while **Fusion** denotes the uncertainty-guided coordinate fusion. The results are reported as average localization error in [m/°].

| SCR+Conf | Fusion | Oxford | QEOxford | NCLT |
|---|---|---|---|---|
|  |  | 2.67/1.25 | 0.83m/1.12 | 1.46/2.80 |
| ✓ |  | 2.56/1.09 | 0.74m/1.00 | 1.30/2.41 |
| ✓ | ✓ | 2.55/1.07 | 0.72m/0.99 | 1.05/2.56 |

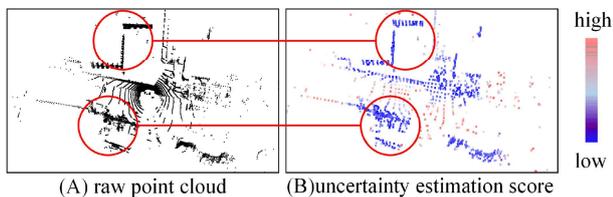

(A) raw point cloud    (B) uncertainty estimation score

Figure 6. **Uncertainty estimation visualization on Oxford.**

hancing localization performance. The bottom row of Fig. 4 shows the localization results for the NCLT dataset. Due to ground truth refinement using SLAM techniques, all methods yield relatively stable localization results. Compared to other methods, our TempLoc achieves more stable localization results. The reduction in outlier predictions further validates the effectiveness of the proposed approach.

**Visualization Comparison.** As shown in Fig. 5. We conducted visualization comparisons on the QEOxford and NCLT datasets. Compared to the state-of-the-art method LightLoc, our network's predicted point clouds in the world coordinate system align more closely with the ground truth.

**Real-time, overhead and performance.** As shown in Tab. 4, we include the representative retrieval-based method BEVplace++ [27] for comparison on NCLT (map: 2012-02-18, query: 2012-05-26) covering real-time performance, computational cost, memory usage, and localization accuracy. As shown, the retrieval-based method BEVPlace++ incurs significantly higher storage overhead than our map-free SCR-based approach. Furthermore, its localization accuracy is substantially lower than our proposed TempLoc.

### 4.3. Ablation Study

**Uncertainty estimation visualization.** As shown in Fig. 6, we visualize the uncertainty estimation. Fig. 6 (A) shows the raw point cloud, and Fig. 6 (B) displays its corresponding uncertainty scores. It can be observed that stable structural elements, such as building walls, exhibit lower uncertainty scores. In contrast, transient or less informative regions like road surfaces and noise have higher uncertainty. Consequently, the uncertainty module guides the network to focus on reliable structural features, thereby enhancing overall localization performance.

**Ablation Study on Different Modules.** To verify each module, we conduct ablation studies on Oxford, QEOxford, and NCLT datasets (Tab. 5). The baseline achieves mean localization errors of $2.67m$, $0.83m$, and $1.46m$, respectively. After integrating the Uncertainty Estimation module (SCR+Conf), translation and rotation accuracy improve by $4.1\%/10.8\%/10.9\%$ and $12.8\%/10.7\%/13.9\%$ across the three datasets, respectively. The gain on Oxford is limited due to ground-truth errors. With the Fusion module, improvements in translation on Oxford and QEOxford are marginal because of their stable environmental structures. However, in the complex campus environment of NCLT, the performance boost over SCR+Conf is significant, with translation error decreasing by $19.2\%$. These results demonstrate that TempLoc effectively leverages temporal relationships and uncertainty estimation to enhance overall localization performance.

## 5. Conclusion

In this paper, we present a new LiDAR relocalization framework, TempLoc, which addresses the robustness issue of traditional single-frame methods in dynamic or ambiguous scenarios by explicitly modeling sequential consistency. Following the measurement, prediction, and fusion paradigm, the proposed method achieves robust and accurate 6-DoF pose estimation without relying on dense map storage. Extensive experiments on the Oxford and NCLT datasets demonstrate that TempLoc comprehensively surpasses existing SOTA methods in both translation and rotation accuracy. More critically, TempLoc achieves the transition from relying solely on instantaneous noisy observations to effectively integrating both past and current information.